\documentclass[letterpaper, 10 pt, conference]{ieeeconf}  

\IEEEoverridecommandlockouts                              

\usepackage{resizegather}
\usepackage{enumerate} 
\usepackage{times}
\usepackage{epsfig}
\usepackage{graphicx}
\usepackage{amssymb} 
\usepackage{amsmath}
\usepackage[linesnumbered,ruled,vlined]{algorithm2e}
\usepackage{algorithmic}
\usepackage{array}
\usepackage{rotating}
\usepackage{multirow}
\usepackage{csquotes}
\usepackage{stfloats}
\usepackage{color}
\usepackage[dvipsnames]{xcolor}
\usepackage{float}  
\usepackage{etoolbox}
\usepackage{balance}
\makeatletter
\patchcmd{\@makecaption}
  {\scshape}
  {}
  {}
  {}
\makeatletter
\patchcmd{\@makecaption}
  {\\}
  {.\ }
  {}
  {}
\makeatother
\def\tablename{Table}

\SetKwInput{KwInput}{Input}
\newcolumntype{P}[1]{>{\centering\arraybackslash}p{#1}}

\newcommand{\thickhline}{%
    \noalign {\ifnum 0=`}\fi \hrule height 1pt
    \futurelet \reserved@a \@xhline
}
\title{\LARGE \bf
Distributed Multi-Target Tracking in Camera Networks
}

\author{Sara Casao \hspace{0.5cm} 
Abel Naya \hspace{0.5cm} 
Ana C.~Murillo\hspace{0.5cm}
 Eduardo Montijano
\thanks{All authors are with the RoPeRt group, at DIIS - I3A, Universidad de Zaragoza, Spain. {\tt\small \{scasao, abeln, acm, emonti\}@unizar.es}
}
\thanks{
This work has been supported by the ONR Global
grant N62909-19-1-2027 and the Spanish projects PGC2018-098817-A-I00 and PGC2018-098719-B-I00 and RTC-2017-6421-7 (MCIU/AEI/FEDER, UE) and DGA T04-FSE.
}
}

\begin{document}

\maketitle
\thispagestyle{empty}
\pagestyle{empty}

\begin{abstract}
Most recent works on multi-target tracking with multiple cameras focus on centralized systems. In contrast, this paper presents a multi-target tracking approach implemented in a distributed camera network. The advantages of distributed systems lie in lighter communication management, greater robustness to failures and local decision making. On the other hand, data association and information fusion are more challenging than in a centralized setup, mostly due to the lack of global and complete information. The proposed algorithm boosts the benefits of the Distributed-Consensus Kalman Filter with the support of a re-identification network and a distributed tracker manager module to facilitate consistent information.
These techniques complement each other and facilitate the cross-camera data association in a simple and effective manner. 
We evaluate the whole system with known public data sets under different conditions demonstrating the advantages of combining all the modules.
In addition, we compare our algorithm to some existing centralized tracking methods, outperforming their behavior in terms of accuracy and bandwidth usage.
\end{abstract}

\section{Introduction}
\label{sec:intro}

Multi-target tracking systems have a broad range of applications, such as security surveillance or crowd behavior analysis \cite{robin2016multiRobot, ardo2019drone} and there is an increasing effort in the research community to deploy them into real world scenarios~\cite{dendorfer2020motchallenge, chavdarova2018wildtrack}. 
Approaching these problems with multi-camera setups brings additional features and benefits with respect to single camera systems, such as more complete information for large spaces or crowded scenes.
Besides, multi-camera systems naturally lead to the development of distributed solutions, which are lighter in communication demands, more robust to failures and faster in processing time than centralized ones. 

However, the implementation of a multi-target multi-camera tracking in a distributed setup has several challenges. First, the construction of robust and efficient distributed systems in terms of bandwidth usage, consensus between nodes and selection of accurate information to share. 
Second, in the presence of multiple targets, each camera must solve independently a data association problem between measurements and trackers. In contrast with centralized systems, where the central node unifies the high level information returning it labeled to the nodes, in a distributed setup the data association across cameras can only be performed locally  and with partial information. 

\begin{figure}
\begin{center}
\includegraphics[width=0.5\textwidth]{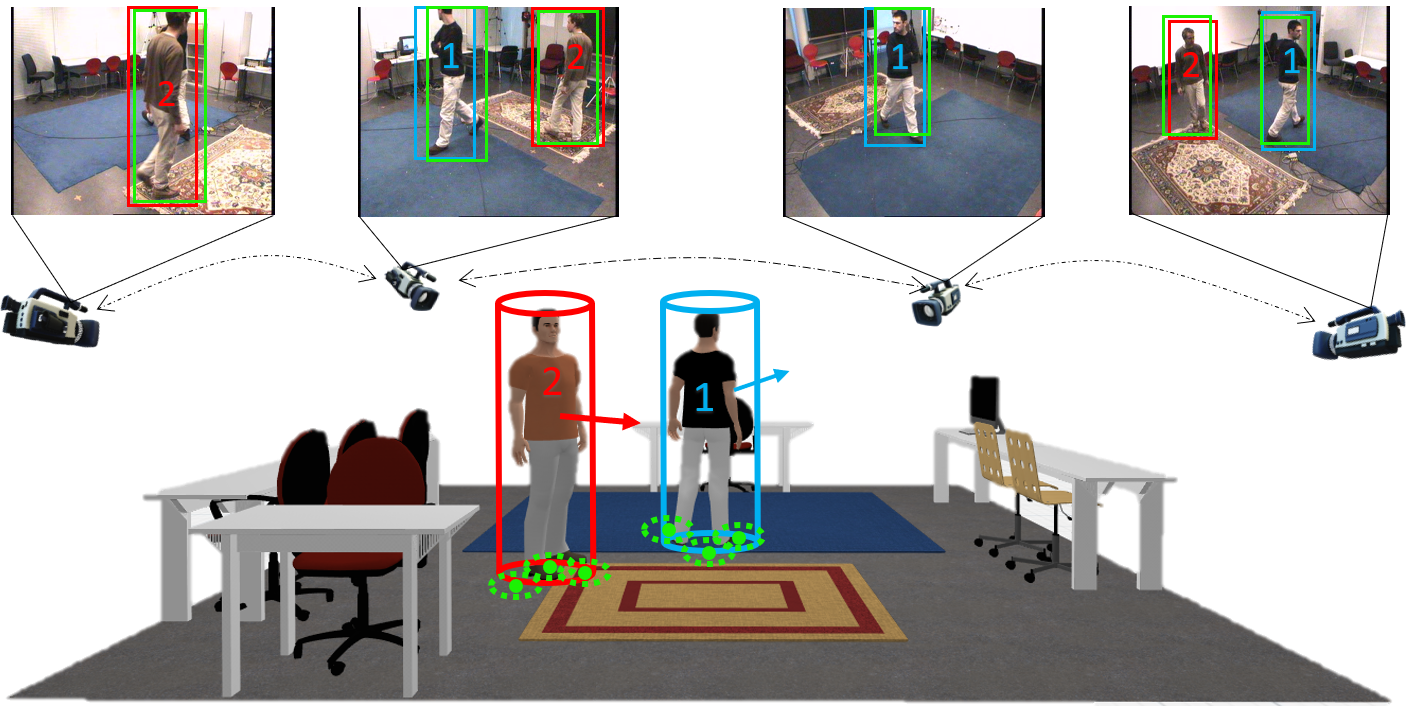}
\end{center}
  \caption{Distributed multi-target tracking scenario. The proposed system maintains in each node the 3D tracking information of all targets (seen or not by the current camera) thanks to the information shared among connected cameras. Each camera solves an independent data association problem. Green boxes correspond to detections while blue and red ones refer to the two existing trackers in the scene, ID1 and ID2 respectively (Best viewed in color).}
\label{fig:globalIdea}
\end{figure}

This work tackles the challenges discussed above by  exploiting synergies between complementary modules typically studied independently in prior work. 
We present an approach for multi-target tracking in a distributed camera network boosting the implementation of a Distributed Kalman Filter (DKF), a local data association process that uses a state of the art re-identification network and a novel distributed method for high level information management.
The overall idea is illustrated in Figure~\ref{fig:globalIdea}. 
The DKF exploits the local data association to decide which information needs to be merged into the distributed consensus update.
Similarly, the data association is enhanced with the uncertainty estimations given by the DKF.
Finally, the tracker manager makes sure of correct association during the initialization and synchronized deletion of the lost trackers.
The main trade-off to consider in our system is bandwidth usage vs accuracy, which we also handle with a careful design.

Specifically, our main contributions are:

$\bullet$ A \textbf{DKF implementation augmented with fully automatic data association} based on geometric cues and appearance information. 
Differently from existing integrated approaches, we run a single communication message per estimation cycle, lightening the process and increasing applicability to real world setups.

$\bullet$ A \textbf{novel distributed strategy to manage the trackers' information}. In contrast with centralized systems, where the central node unifies the information, we propose a distributed data association across cameras that manages the partial information in local nodes. To minimize the excess of bandwidth usage caused by the exchange of appearance information, our algorithm communicates appearance features only once per tracker. 

The proposed approach is evaluated in public benchmarks, demonstrating the benefits with respect to a naive DKF implementation, as well as established centralized algorithms. Experimentation detailed in Section \ref{sec:experiment} analyzes different relevant aspects including
the effect of connectivity restrictions, 
the influence of appearance in the filter 
and the effects of our strategy to unify high level information. 

\section{State of the art}
\label{sec:stateofart}
This section summarizes related work on core multi-target multi-camera tracking aspects.

\subsection{Multi-view multi-target tracking in centralized systems}
Multi-target multi-camera tracking in centralized systems sends the information from each camera to a common location where all the data is processed together  \cite{tesfaye2019fastconstrains}. These implementations, which stand out for their accuracy, are typically used in safety applications \cite{ferraguti2020safety1,chen2018safety2}.
Obtaining the complete trajectories of several targets is normally formulated as an optimization problem in a graph. The nodes represent short trajectories, known as tracklets, obtained by the association of detected bounding boxes. There are different variations to compute the weights of the graph. The combination of the similarity measure given by a triplet loss with a linear motion model is proposed in~\cite{ristani2018features}. The proposal in \cite{wen2017hyper-graph} is the association of tracklets across views based on correlations in motion, appearance and smoothness of the resulting 3D trajectory. A method is presented in \cite{le2018online} to select a subset of tracklets from the graph and associate them based on geometry and motion cues. Other works such as \cite{xu2016hierarchy} model the tracklet association problem as a hierarchical structure optimization.
The main disadvantages of centralized methods are the excessive bandwidth usage, required to send all the information to the central computer, and the lack of robustness with respect to a single point of failure.

\subsection{Multi-view multi-target tracking in distributed systems}
Distributed implementations typically focus on improving the robustness and the efficiency through the study of the bandwidth, the consensus between nodes and the accuracy of the information shared. Several algorithms are proposed in \cite{olfati2007DKF} to achieve a consensus in a distributed heterogeneous sensor network performing only one communication per estimation cycle. One of those algorithms is implemented in \cite{soto2009distributed} for a Pan-Tilt-Zoom (PTZ) camera network to track people of interest, although the data association problem is not addressed there. The work in \cite{kamal2015ICF&JPDAF} selects the Information-weighted Consensus Filter (ICF) method \cite{kamal2012ICF} as consensus algorithm and fills the gap of data association with the Joint Probabilistic Data Association (JPDAF) algorithm \cite{zhou1993JPDAF}, which uses the previous target states to relate measurements and trackers. A drawback of the ICF method is the use of several communication messages per camera and estimation cycle, being less efficient and partially breaking the appealing properties of pure distributed systems. Using the same consensus algorithm, \cite{he2019efficient} proposes a tracking approach in a distributed camera network. They address data association within each camera through a global metric, merging appearance and geometry cues, and across-view data association through the euclidean distance between the 3D position of the targets. The improvement of the Consensus Kalman Filter is discussed in \cite{shorinwa2020distributedVehicles} posed as a Maximum A Posteriori (MAP) optimization problem. The algorithm consists of closed-form algebraic iterations that guarantee the convergence to the centralized MAP over a designated sliding time window. They assume that each target in the environment has a unique identifier known to all sensors.
Our multi-target multi-camera tracking approach is based on \cite{soto2009distributed}, and fills the gap of automatic data association between measurements and trackers with a global metric based on geometry and appearance. Unlike the consensus algorithm in \cite{he2019efficient}, the algorithm implemented in this work only sends one communication message per estimation cycle, lightening the communication process. Besides, we include a specific strategy to manage the high level data across cameras to improve the global data association and the consistency of the trackers in the network. 

\subsection{Data association and re-identification}
A wide variety of techniques have been proposed to tackle the problem of data association.
One of the most popular techniques is the use of motion models to compute similarity based on geometric constraints, and the use of features such as histograms for appearance criteria \cite{de2020divers,chandra2019densepeds}. Commonly, a global similarity function is defined based on both metrics. Other works extract body poses and relate them by the nearest observation statistically consistent with the distribution of positions \cite{virgona2018shoulder}. Taking into account the factors included in the data association problem, previous work defines several costs related to geometry, shape, appearance, pose and coordinate transformation to obtain a complete similarity function \cite{sharma2018cost}. The work proposed in \cite{tsai2019intertial} uses inertial sensing and RGB-D cameras to capture the skeleton data and perform a short-term pairing. A long-term pairing process adds the color histogram to the similarity function in order to increase robustness. The data association process in our approach is similar to \cite{de2020divers}, but our implementation takes advantage of re-identification strategies supported by recent deep learning techniques. A generalized strategy is based on comparing feature vectors, obtained from a network output, to measure the similarity between a query image and a global gallery (i.e., model) of individuals \cite{chen2019abd,quan2019auto,liu2018attributerecognition}. In order to be efficient and effective, to extract these appearance feature vectors we use the architecture proposed in \cite{zhou2019omni}, that mixes global and local features in a lightweight network, changing traditional convolution operations by depth-wise operations. 

\section{Distributed Tracking Approach}
\label{sec:method}

\begin{figure}
\begin{center}
\includegraphics[width=0.5\textwidth]{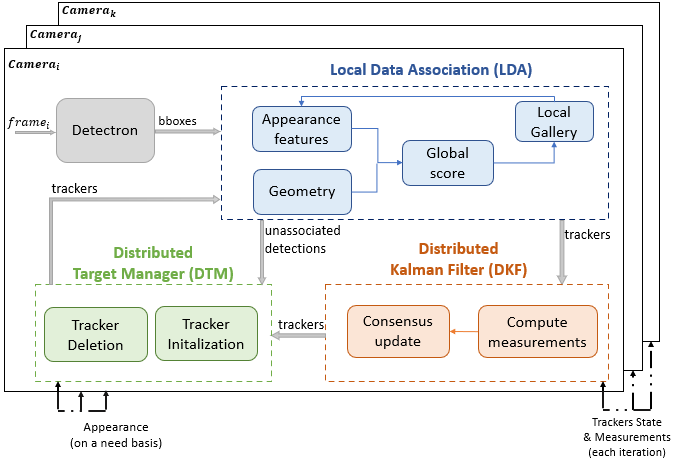}
\end{center}
   \caption{Overall architecture of our distributed multi-target multi-camera tracking system.}
\label{fig:arquitecture}
\end{figure}
Figure \ref{fig:arquitecture} summarizes the proposed architecture. 
First, image target positions are obtained with a people detector~\cite{wu2019detectron2} in each camera. Then, the association between the current detections and existing trackers is performed locally (LDA). This association is based on a global score computed from the geometric cues, provided by the local tracking filter, and the appearance similarity with respect to the appearance model of the target stored in the local gallery. To continue the cycle, each camera exchanges a single communication message with the neighboring cameras, the trackers get into the DKF where the new state of the target is updated and sent to the Distributed Target Manager (DTM) block. Finally, the DTM manages the initialization of new trackers and the deletion of unobserved ones uniformly over the network.

\subsection{Distributed Kalman Filter}
For simplicity in the exposition, throughout this sub-section we will consider the distributed tracking of a single target.
The target model, \sloppy $\mathbf{x}(k) = (x(k), y(k), w(k), h(k), \dot{x}(k), \dot{y}(k)),$ is represented as a 3D cylinder moving on a ground plane, where  $(x(k), y(k))$ are the ground plane coordinates of the cylinder center, $( w(k), h(k))$ are the width and height of the cylinder and $(\dot{x}(k) , \dot{y}(k))$ are the velocity of the target in the $x$ and $y$ directions.
The filter models the motion of the target considering a discrete-time linear dynamical system, 
\begin{gather}
\mathbf{x}(k+1) = \mathbf{A}\mathbf{x}(k) + \mathbf{w}(k),
\label{transitionDynamics}\hskip 0.25cm 
\mathbf{z}(k) = \mathbf{H}\mathbf{x}(k) + \mathbf{v}(k),
\end{gather}
where $\mathbf{w}(k)$  and $\mathbf{v}(k)$ are zero mean Gaussian noise ($\mathbf{w}(k)\sim \mathcal{N}(0 ,\mathbf{Q}(k)), \mathbf{v}(k)\sim \mathcal{N}(0,\mathbf{R}(k))$), being $\mathbf{Q}(k)$ and $\mathbf{R}(k)$ the model and measurement covariance matrices respectively. 
The change of the target model in each step depends directly on the transition matrix $\mathbf{A}$, which considers a constant velocity model on the target position and no dynamics on the rest of the state. 
The noisy measurement $\mathbf{z}(k)$ is the 3D cylinder obtained as the projection of the bounding box given by the detector, i.e., $\mathbf{z}(k)=(x(k), y(k), w(k), h(k))$ defined with the output matrix $\mathbf{H}$ plus the noise.
To obtain this projection we assume known homographies for each camera to map the image plane to ground plane coordinates. 

The independent execution of the filter in each camera, $C_i,$ produces a local estimation of the target, $\hat{\mathbf{x}}_i(k)$, possibly different to other cameras' estimation. A Distributed Kalman-Consensus filter is used to mitigate these differences. The consensus algorithm works with known data association between the local measurement, $\mathbf{z}_i(k)$, and target prediction for all the cameras. 
From this association, each camera computes its sensor data information, $\mathbf{u}_i(k)$, and its inverse-covariance matrix, $\mathbf{U}_i(k)$, defined as the vector and matrix information, obtained as
\begin{gather}
\mathbf{u}_i(k) = {\mathbf{H}}^T{\mathbf{R}_i}^{-1}(k)\mathbf{z}_i(k), \hspace{2mm} \mathbf{U}_i(k) = {\mathbf{H}}^T{\mathbf{R}_i}^{-1}(k)\mathbf{H}. \label{matInfo}
\end{gather}
The values in~\eqref{matInfo} are exchanged with neighboring cameras in the network, $C_j\in C_i^n$, together with the prediction of the target state, $\mathbf{\bar{x}}_i(k)$, obtained as $\mathbf{\bar{x}}_i(k) =\mathbf{A}\mathbf{\hat{x}}_i(k-1)$.
Assuming the measurement noises of the sensors are uncorrelated, the representation in information form allows the cameras to combine all the received measurements with the acquired one by simply adding them,
\begin{gather}
\mathbf{y}_i(k) = \displaystyle\sum_{C_j\in C_i^n} \mathbf{u}_j(k), \hspace{5mm}
\mathbf{S}_i(k) = \displaystyle\sum_{C_j\in C_i^n} \mathbf{U}_j(k).
\label{calcS}
\end{gather}
The state is then updated by the correction in the prediction of the target state with the merged information and the predictions from the neighboring cameras, 
\begin{equation}
    \begin{split}
\mathbf{\hat{x}}_i(k) = \mathbf{\bar{x}}_i(k) &+ \mathbf{M}_i(k)\left[\mathbf{y}_i(k) - \mathbf{S}_i(k)\mathbf{\bar{x}}_i(k)\right] \\ &+\gamma \mathbf{M}_i(k)\displaystyle\sum_{C_j\in C_i^n}(\mathbf{\bar{x}}_j(k)-\mathbf{\bar{x}}_i(k)), \label{xhat}
    \end{split}
\end{equation}
where $\mathbf{M}_i(k) = (\mathbf{P}_i(k)^{-1} + \mathbf{S}_i(k))^{-1}$ is the Kalman Gain in the information form, $\mathbf{P}_i(k)$ is the covariance of the target state and $\gamma = 1/\|\mathbf{M}_i(k) + 1\|$.
Finally, the covariance matrix is updated according to $\mathbf{P}_i(k+1) = \mathbf{A}\mathbf{M}_i(k){\mathbf{A}}^T+ \mathbf{Q}_i(k) \label{P}.$

Although the standard implementation of the DKF assumes synchronization of the cameras, the consensus-nature of the algorithm makes it amenable to a fully asynchronous, event-triggered implementation such as~\cite{he2017event}. 

\subsection{Local Data Association}
\label{sec:lda}
In our method, the local data association required for a correct update of the filter is made merging two constraints based on geometry and appearance.
Let us consider now that in a particular estimation cycle, the set of measurements $\mathcal{Z}=\{\mathbf{z}_j\}$\footnote{We use now the index $j$ to denote different measurements observed in a single camera instead of neighbors in the camera network.} is provided by the detector to the LDA module.
Since the DKF updates the uncertainty of the tracker state, we take advantage of this information to calculate the Mahalanobis distance between the $x,y$ position on the ground of each measurement, $\mathbf{z}_j$, and the predicted position, $\mathbf{\bar{x}}_i$,
\begin{equation}
d(\mathbf{z}_j, \mathbf{\bar{x}}_i) = \sqrt{(\mathbf{z}_j-\mathbf{H\bar{x}}_i)\mathbf{V}^{-1}(\mathbf{z}_j-\mathbf{H\bar{x}}_i)^T}, 
\label{eq:maha}
\end{equation}
being $\mathbf{V} = \mathbf{P}_{xy}+\mathbf{R}_{xy}$, with $\mathbf{P}_{xy}$ and $\mathbf{R}_{xy}$ the sub-matrices of $\mathbf{P}_i$ and $\mathbf{R}_j$ that encode the position covariance of the estimation and the measurement respectively. 
Then, the similarity value in geometry is computed as
\begin{align}
s_d(\mathbf{z}_j, \mathbf{\bar{x}}_i) = \left\{ \begin{array}{c} 
                \frac{1}{\alpha}\hspace{1mm}d(\mathbf{z}_j, \mathbf{\bar{x}}_i) \hspace{3mm} \hbox{ if } \hspace{1.5mm}d(\mathbf{z}_j, \mathbf{\bar{x}}_i) < \tau \\
                1 \hspace{10mm} \hbox{ otherwise},             \\
                \end{array} \right.
\label{eq:sd}
\end{align}
where $\alpha$ is a configuration parameter and $\tau$ a threshold applied to ignore highly unlikely candidates.

The candidates selected by the geometry constraint are then evaluated in appearance. Instead of using traditional hand-crafted descriptors to measure the appearance similarity, we employ a network designed for people re-identification \cite{zhou2019omni}, whose weights have been pre-trained with the MSMT17 Benchmark \cite{wei2018msmt17}. 
Inspired by the re-identification task evaluation methodology, we associate with each tracker a local gallery of limited size that represents the target appearance model. 
The construction of this gallery is currently done by simply storing locally observed patches for each target and updating them over time, saving a new patch every $N$ frames, and discarding the oldest one.
The appearance similarity between a query and the tracker gallery is obtained with the minimum cosine distance,
\begin{equation}
s_a(\mathbf{a}_j) = \min_{\mathbf{a}_{i} \in \mathcal{A}_{i}}
\displaystyle\left( 1 - \frac{\mathbf{a}_j^T \hspace{1mm} \mathbf{a}_{i}}{\|\mathbf{a}_j\| \|\mathbf{a}_{i}\|}\right),
\label{eq:sa}
\end{equation}
where $\mathbf{a}_j,$ the query, is the appearance feature vector associated to detection $\mathbf{z}_j$, and $\mathcal{A}_i=\{\mathbf{a}_i\}$ is the set of feature vectors of the target's gallery used by camera $C_i$. The final decision in the data association is based on selecting the minimum value of the product of both scores, $s_a$ and $s_d$.

\subsection{Distributed Tracker Manager}
\label{sec:manage}
Another important issue to address in a practical implementation of the DKF is the management of the trackers through the full distributed system. This requires a correct data association of trackers across different cameras to guarantee that the information mixed in~\eqref{calcS} and~\eqref{xhat} corresponds to the same target. 
Similarly, the cameras need to agree upon the time in which a particular tracker is no longer relevant and should be dropped.
We propose how to address these problems in a distributed fashion.
\vspace{2mm}
\paragraph*{Distributed Global Data Association}
Trackers are identified locally by a two-dimensional unique identifier, ID$_i$, described by the camera id in the network, $i$, and a local counter, $n$.
New trackers can either be initialized because of a new local observation or because of a transmission from neighboring cameras.
Local initialization is done whenever a new target generates two observations in its local gallery. This helps filtering spurious measurements from the detector, giving enough time for a new tracker to ensure that it corresponds to a valid target.
Once this happens, we attach to the DKF data the appearance model, $\mathcal{A}_{\hbox{ID}_i},$ in the message sent to neighbors.
It is important to highlight that this is the only moment when appearance is transmitted through the network in our algorithm, consisting in the two descriptors available in the local gallery at that time.
The second case that can trigger new tracker initializations in our system is the reception of  messages from neighbor cameras. Our algorithm considers three situations for this case:
1) A single neighbor camera sends a new tracker.
Then, the camera creates a new tracker and associates it to the received one for the future DKF consensus updates. The local gallery is initialized with the appearance model received.
2) The camera receives new trackers from several neighboring cameras.
3) A new local tracker is initialized at the same time that new trackers from other cameras are received.
In situations 2) and 3), it is necessary to check whether the new trackers from the different cameras are of the same target or not. 
We perform a similar process to the one for local data association described in Section \ref{sec:lda}, replacing in~\eqref{eq:maha} the measurement by the other camera's estimation, $d(\mathbf{\hat{x}}_i, \mathbf{\bar{x}}_j)$, and the Mahalanobis distance with the Euclidean distance, $\mathbf{V}=\mathbf{I}$, since the covariance matrices associated to the trackers are not part of the communication messages. This also requires a different threshold in~\eqref{eq:sd}. If two or more trackers are similar enough, they are merged locally into a single one.
\vspace{2mm}
\paragraph*{Consensus-based Tracker Drop and Re-initialization} The other main task of the Distributed Tracker Manager is to decide when to drop a tracker. Instead of letting each camera to decide this process individually, we have opted for a consensus-based solution that reduces, as much as possible, the number of iterations that different cameras carry out the tracking individually.
We let $\ell_i$ be the local estimation that camera $i$ has on the number of iterations gone since the last local data association of a measurement to the target made by any camera of the network. Since this is a global parameter that involves the whole network, its estimation is sent as part of the tracker message at every iteration.
If the camera achieves the local data association of the tracker,
the new value of this parameter is set to zero. Otherwise, the camera chooses as new value the minimum among all the values received, including its own, and adds one unit,
\begin{equation}
 \ell_{i}(k+1) =
 \left \{
 \begin{array}{ll}
   0 & \textrm{if detected} \\
   \displaystyle\min_{j \in \mathcal{C}_{i}^n}
\left(\ell_i(k), \ell_j(k)\right)+1 & \textrm{otherwise}
 \end{array}
\right.
\label{eq:ages}
\end{equation}

The camera drops the target when $\ell_i(k+1)$ is higher than a threshold $\kappa$. The tracker ID is saved together with its gallery as an \textit{old tracker}, to be recovered if the same target is back in the camera field of view or some other camera re-activates it. This process is checked during the local initialization. Before assigning a new ID, a re-identification score is computed between the new tracker gallery and the galleries from \textit{old trackers} through~\eqref{eq:sa}. The re-initialization is accomplished if $s_a$ is lower than $\epsilon$.

\subsection{Communication and bandwidth requirements.}
\label{Sec:BW}
The information shared between cameras at every iteration consists, for each active tracker, of the tracker ID, its predicted state, the measurements obtained in~\eqref{matInfo} and the counter of the last observation, (ID$_i$, $\mathbf{\bar{x}_i}, \mathbf{u}_i, \mathbf{U}_i$, $\ell_i$).
This message is encoded using a total of $51$ elements, representing less than $1$kB of bandwidth information per tracker.
In order to carry out the process explained in the \textit{Distributed Global Data Association}, the message with the information related to the new trackers is sent together with appearance information. The appearance information exchange between cameras is composed by two appearance feature vectors required in the trackers initialization. Each one has 512 elements, representing $16.38$kB of bandwidth information that is sent only once per tracker.  
\section{Experiments}
\label{sec:experiment}
\subsection{Experimental setup.}
Our experiments are run on data from the EPFL dataset~\cite{k-short}: \textit{Terrace} set with 4 outdoor cameras, \textit{Laboratory}, with 4 indoor cameras and \textit{Campus} set with 3 outdoor cameras. 
For the person detection, all experiments run the official Detectron implementation \cite{wu2019detectron2}, filtering the output with a process equivalent to non-maximal suppression.
Regarding evaluation, we use standard Multiple Object Tracking (MOT) metrics, including Multi Object Tracking Accuracy (\textbf{MOTA}) and Multiple Object Tracking Precision (\textbf{MOTP}), as defined in \cite{bernardin2008evaluating}, and Identification Precision (\textbf{IDP}), Identification Recall (\textbf{IDR}) and their F1 Score (\textbf{IDF1}) as defined in \cite{ristani2016evaluating2}. 
All the evaluations follow the same process explained in \cite{xu2016hierarchy}, giving as a final result the median of each metric for all the cameras available in each dataset.
\begin{center}
\begin{figure*}[tb!]
    \begin{tabular}{@{}ccccc}
    \begin{tabular*}{13mm}[l]{@{}c}
    \end{tabular*} &
    \begin{tabular}{cccc}
    Complete \hspace{26mm} & Ring \hspace{29mm} & Chain \hspace{20mm} & Disconnected
    \end{tabular}\\
    \begin{tabular*}{13mm}[l]{@{}c}
    Graph
    \end{tabular*} & 
    \begin{tabular}{cccc}
    \includegraphics[width=12mm, height=12mm]{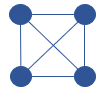} \hspace{23mm} &
    \includegraphics[width=12mm, height=12mm]{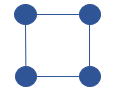} \hspace{25mm} &
    \includegraphics[width=12mm, height=12mm]{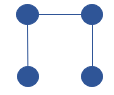} \hspace{22mm} &
    \includegraphics[width=12mm, height=12mm]{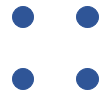} 
    \end{tabular}\\
    \begin{tabular*}{13mm}[l]{@{}c}
    Terrace
    \end{tabular*} &
    \begin{tabular}{cccc}
    \includegraphics[width=37mm, height=27mm]{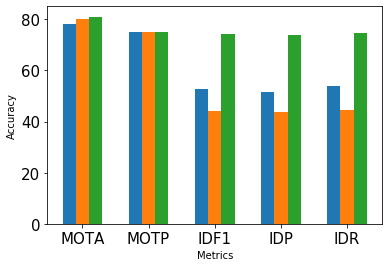} &
    \includegraphics[width=37mm, height=27mm]{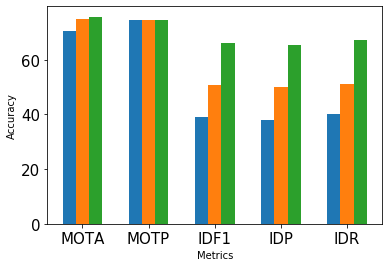} & 
    \includegraphics[width=37mm, height=27mm]{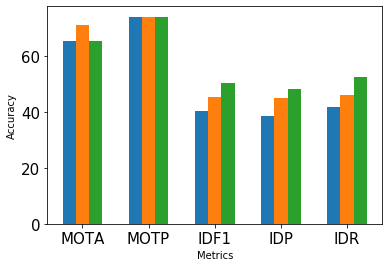} &
    \includegraphics[width=37mm, height=27mm]{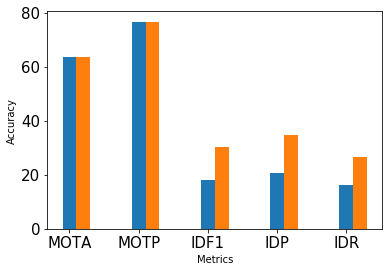}
    \end{tabular}\\
    \begin{tabular*}{13mm}[l]{@{}c}
    Laboratory
    \end{tabular*}&
    \begin{tabular}{cccc}
    \includegraphics[width=37mm, height=27mm]{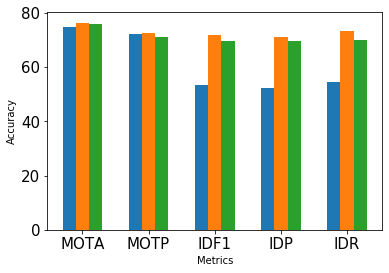} & 
    \includegraphics[width=37mm, height=27mm]{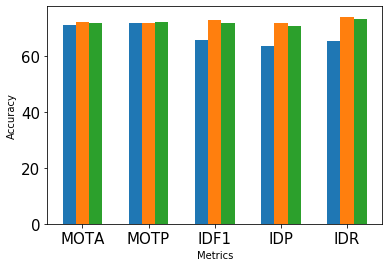} & 
    \includegraphics[width=37mm, height=27mm]{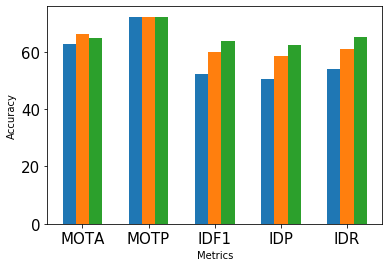} &
    \includegraphics[width=37mm, height=27mm]{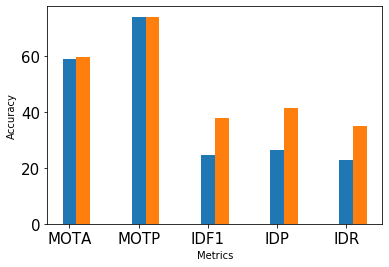}
    \end{tabular} \\
    \begin{tabular*}{13mm}[l]{@{}c}
    Campus*
    \end{tabular*} &
    \begin{tabular}{cccc}
    \includegraphics[width=37mm, height=27mm]{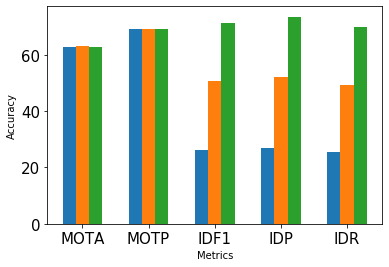} & 
    \includegraphics[width=37mm, height=27mm]{images/Campus_Ring.png} & 
    \includegraphics[width=37mm, height=27mm]{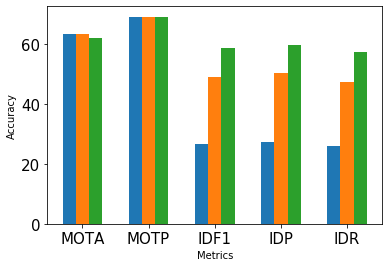} &
    \includegraphics[width=37mm, height=27mm]{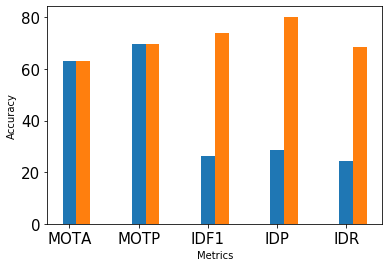}
    \end{tabular}\\
    \multicolumn{4}{c}{\footnotesize \hspace{65mm} \includegraphics[width=54mm, height=2.8mm]{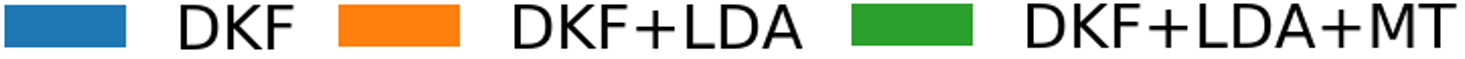}} \\
    \end{tabular}
     \caption{Ablation study for our approach considering different graph topologies (complete, ring, chain, disconnected). The variations are run on three sets of data (Terrace, Laboratory and Campus) and evaluated using CLEAR MOT metrics. Running all the steps of the proposed approach (DKF+LDA+MT) achieves the best results in the majority of cases. *The Campus dataset only has three cameras, obtaining the same results in the complete and ring graphs.}
    \label{fig:histograms}
\end{figure*}
\end{center}
\vspace{-6mm}
\subsection{Ablation Study and Topology effect}
This first experiment analyzes three configurations of the proposed approach to evaluate the effects of the novel modules we propose in our architecture. The simplest configuration is our implementation of the Distributed Kalman Filter using only geometric information in the data association process (DKF). A second version adds the appearance information from the local gallery (DKF + LDA). The latest version includes the distributed tracker management module in the system. This version represents the complete algorithm presented in the paper (DKF + LDA + DTM). 

\begin{figure}[H]
\centering
\includegraphics[width=0.45\textwidth]{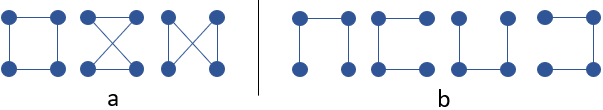}
\caption{Alternatives considered for (a) ring graph topologies and (b) chain graph topologies.}
\label{fig:connections}
\end{figure}

Besides the ablation study, this experiment analyzes the effects of different network topologies. For chain and ring topologies we have considered several network alternatives to make the evaluation independent of the individual quality of particular cameras in the tracking. The considered combinations are shown in Figure \ref{fig:connections}. Furthermore, we evaluate the influence of the appearance in a disconnected graph, i.e., without information shared between cameras and thus running four independent Kalman filters.
Figure \ref{fig:histograms} summarizes the results of all the tests in this experiment.  Each row shows the results obtained using certain dataset, and each column corresponds to the results with a different connectivity graph. 
In all of them, the configuration parameters from the proposed algorithm are set to $\alpha_{LDA}=2000$, $\alpha_{GDA}=50$, $\tau=0.5$, $\kappa=15$, $\epsilon = 0.25$ and the gallery size to 20 samples with $N=20.$ 
These parameters were defined in Sections~\ref{sec:lda} and~\ref{sec:manage}.
\begin{figure}[!tb]
\centering
\includegraphics[width=0.48\textwidth]{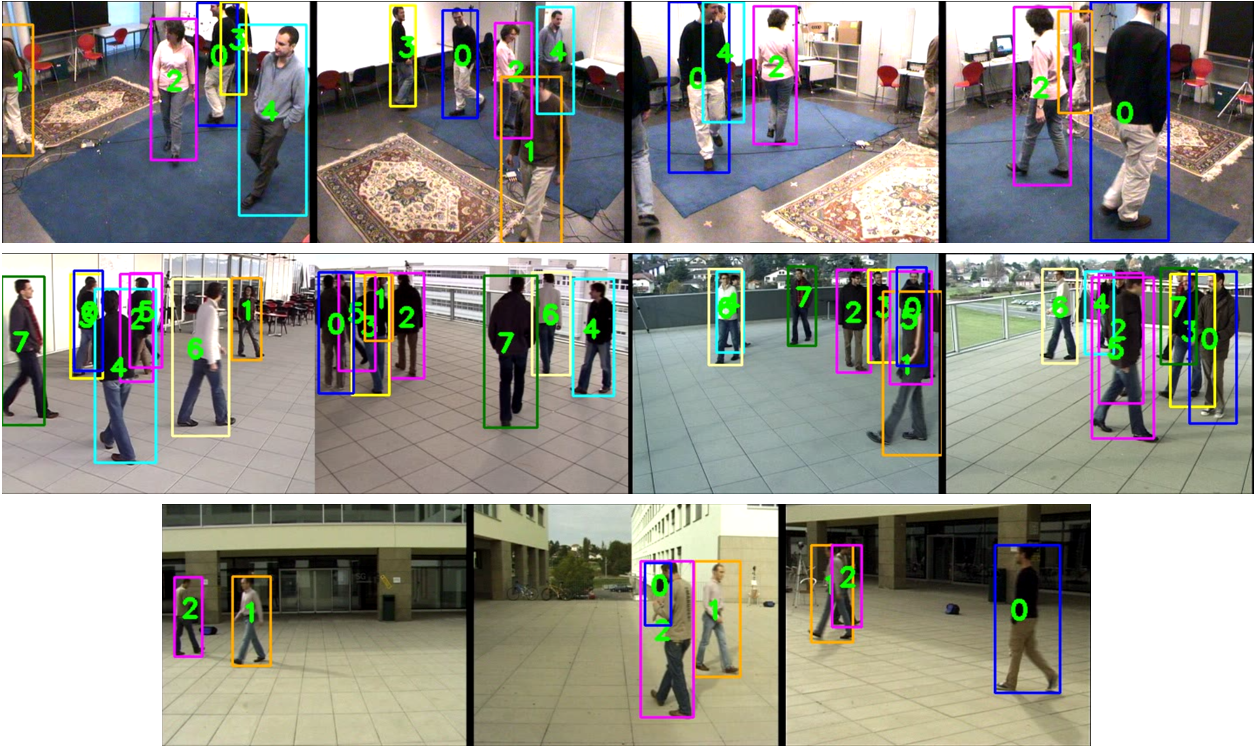}
\caption{Qualitative tracking results of our approach in three datasets: \textit{Laboratory} (first row), \textit{Terrace} (second row) and \textit{Campus} (third row). More on the supplementary video.}
\label{fig:epfl}
\end{figure}
\\Overall, we can see how both proposed modules (LDA and DTM) bring significant improvements in all the metrics and datasets, with more positive influence in the identification metrics (IDF1, IDP and IDR) than the tracking metrics (MOTA, MOTP). 
For a few cases it would be slightly better to apply only LDA, but the penalization for including DTM in those few cases (decrease of less than 5\%) is not nearly as significant as the benefits it brings in the rest of the cases (up to 20\% increase in some network topologies with respect to the DKF+LDA configuration). 
Figure \ref{fig:epfl} shows some qualitative results for each of the datasets. The supplementary material provides additional results.  

\subsection{Comparison with other algorithms}
This experiment compares the proposed approach (configured exactly as in the previous experiment) with recent centralized methods evaluated in \cite{xu2016hierarchy}. 
They publish results on public datasets running their proposed approach, Hierarchical Trajectory Composition (HTC), as well as two other baselines, Probabilistic Occupancy Map (POM) \cite{pom} and K-shorted Path (KSP) \cite{k-short}. 
To make a fair comparison, the evaluation parameters are those described in \cite{xu2016hierarchy}. Among the datasets used there that provide accurate camera-ground plane calibration (required to run our algorithm), we picked the most challenging sequence of \textit{Terrace}, where a high number of crossings and occlusions occur between targets. Note this does not intend to be a thorough evaluation, but rather an experiment to see where the proposed distributed approach gets in comparison with existing centralized approaches.

Previously discussed distributed approaches \cite{soto2009distributed, kamal2015ICF&JPDAF} can not be included in this study because up to our knowledge they do not provide MOT metric results in available benchmarks, or focus on different goals and metrics than us. 
In \cite{kamal2015ICF&JPDAF}, the experiments focused on the consensus algorithm analysis, whereas \cite{soto2009distributed} uses its own camera network to test the proposed approach, showing as result the trajectories of the individuals on the ground.
Section~\ref{sec:stateofart} already highlighted our approach advantages with respect to these systems.

\begin{table}[!tb]
\centering
\footnotesize
\begin{tabular}{|p{3.9cm}|c|c|c|}
\hline
\textbf{Algorithm} & MOTA & MOTP & Bandwidth\\
  &   &   & [kB/frame]$^{o}$\\
\hline
\multicolumn{4}{|l|}{\textbf{Centralized} baselines}\\
\hline
HTC \cite{xu2016hierarchy}* & 71.84 & 71.15 & 1215 \\
KSP \cite{k-short}*         & 65.75 & 57.82 & 1215 \\
POM \cite{pom}*             & 56.9  & 61.33 & 1215 \\
\hline
\multicolumn{4}{|l|}{\textbf{Distributed} proposed approach}\\
\hline
DKF+LDA+DTM (Complete) & 80.95 & 75.2  & 28.23 \\
DKF+LDA+DTM (Ring) & 75.98 & 74.42 & 28.23 \\
DKF+LDA+DTM (Chain) & 69.74 & 73.96  & 28.23 \\
\hline 
\multicolumn{4}{@{}p{8cm}}{\footnotesize * Results interpolated \cite{xu2016hierarchy}, only shown graphically.}\\
\multicolumn{4}{@{}p{8cm}}{\footnotesize $^{o}$ Bandwidth calculated analytically more details in the text.}
\end{tabular}
\caption{MOT metrics and bandwidth requirements per frame of our distributed approach and existing centralized methods on \textit{Terrace} dataset.}
\label{tab:comparison2}
\end{table}

The results are summarized in Table \ref{tab:comparison2}, where we see better performance of our approach with respect to the centralized methods in two of the three graph topologies. The complete graph, being the closest version to the centralized system, gets an improvement of 9.11\% in MOTA while the ring graph improves the results by 4.14\%. Finally the chain graph obtains lower MOTA, but still comparable results to those of the HTC algorithm.
Regarding the bandwidth, for the centralized systems we have computed it considering that every camera is at one hop communication to the central server. Since there are four cameras, the server needs to receive 4 images of $288\times360$ RGB pixels each one, giving a total of 414,720 pixels. Each pixel is encoded with 3 Bytes (R+G+B), so the total bandwidth required is 1,215kB/frame.
For the distributed system, we compute an upper bound of the bandwidth, considering that every iteration there are $9$ active trackers, which is the maximum number of trackers active during the whole execution.
Each tracker requires $0.78$kB/frame (Section~\ref{Sec:BW}), so when multiplied by the 4 cameras and the 9 trackers results $28.08$kB.
The total number of trackers initialized in our algorithm is $24$, that over all the iterations average a total of $0.15$kB/frame to send the appearance information of the new trackers.
The sum of these two quantities results in the $28.23$kB/frame of Table~\ref{tab:comparison2}.
\balance

\section{Conclusions}
This paper has presented a new multi-target tracking approach for a distributed camera network, providing a global approach that deals with the distributed fusion of low and high level information. 
In this work, the challenges of a distributed system have been addressed boosting the DKF with a fully automatic data association and a novel tracker manager to handle the misalignment of the high level information. 
The data association is based on geometric and appearance constraints. 
The distributed tracker manager takes care of the global data association and each tracker's consistency in the network to reduce the number of iterations that different cameras carry out the tracking individually. 
The proposed approach is evaluated in challenging public benchmarks, 
reaching comparable, or even better results than centralized systems and demonstrating the benefits with respect to a naive DKF implementation.
The current implementation of the DKF assumes synchronization of the cameras but the consensus-nature of the algorithm makes it amenable to a fully asynchronous and event-triggered implementation. This and more sophisticated strategies of information selection to build the galleries are open challenges left for future work.

{
\bibliographystyle{IEEEtran}
\bibliography{egbib}

\begin{thebibliography}{10}
\providecommand{\url}[1]{#1}
\csname url@rmstyle\endcsname
\providecommand{\newblock}{\relax}
\providecommand{\bibinfo}[2]{#2}
\providecommand\BIBentrySTDinterwordspacing{\spaceskip=0pt\relax}
\providecommand\BIBentryALTinterwordstretchfactor{4}
\providecommand\BIBentryALTinterwordspacing{\spaceskip=\fontdimen2\font plus
\BIBentryALTinterwordstretchfactor\fontdimen3\font minus
  \fontdimen4\font\relax}
\providecommand\BIBforeignlanguage[2]{{%
\expandafter\ifx\csname l@#1\endcsname\relax
\typeout{** WARNING: IEEEtran.bst: No hyphenation pattern has been}%
\typeout{** loaded for the language `#1'. Using the pattern for}%
\typeout{** the default language instead.}%
\else
\language=\csname l@#1\endcsname
\fi
#2}}

\bibitem{robin2016multiRobot}
C.~Robin and S.~Lacroix, ``Multi-robot target detection and tracking: taxonomy
  and survey,'' \emph{Autonomous Robots}, vol.~40, no.~4, pp. 729--760, 2016.

\bibitem{ardo2019drone}
H.~Ardo and M.~Nilsson, ``Multi target tracking from drones by learning from
  generalized graph differences,'' in \emph{IEEE Int. Conf. on Computer Vision
  Workshops}, 2019.

\bibitem{dendorfer2020motchallenge}
P.~Dendorfer, A.~Osep, A.~Milan, K.~Schindler, D.~Cremers, I.~Reid, S.~Roth,
  and L.~Leal-Taix{\'e}, ``Motchallenge: A benchmark for single-camera multiple
  target tracking,'' \emph{International Journal of Computer Vision}, pp.
  1--37, 2020.

\bibitem{chavdarova2018wildtrack}
T.~Chavdarova, P.~Baqu{\'e}, S.~Bouquet, A.~Maksai, C.~Jose, T.~Bagautdinov,
  L.~Lettry, P.~Fua, L.~Van~Gool, and F.~Fleuret, ``Wildtrack: A multi-camera
  hd dataset for dense unscripted pedestrian detection,'' in \emph{IEEE Conf.
  on Computer Vision and Pattern Recognition}, 2018, pp. 5030--5039.

\bibitem{tesfaye2019fastconstrains}
Y.~T. Tesfaye, E.~Zemene, A.~Prati, M.~Pelillo, and M.~Shah, ``Multi-target
  tracking in multiple non-overlapping cameras using fast-constrained dominant
  sets,'' \emph{International Journal of Computer Vision}, vol. 127, no.~9, pp.
  1303--1320, 2019.

\bibitem{ferraguti2020safety1}
F.~Ferraguti, C.~T. Landi, S.~Costi, M.~Bonf{\`e}, S.~Farsoni, C.~Secchi, and
  C.~Fantuzzi, ``Safety barrier functions and multi-camera tracking for
  human--robot shared environment,'' \emph{Robotics and Autonomous Systems},
  vol. 124, 2020.

\bibitem{chen2018safety2}
J.-H. Chen and K.-T. Song, ``Collision-free motion planning for human-robot
  collaborative safety under cartesian constraint,'' in \emph{IEEE Int. Conf.
  on Robotics and Automation}, 2018.

\bibitem{ristani2018features}
E.~Ristani and C.~Tomasi, ``Features for multi-target multi-camera tracking and
  re-identification,'' in \emph{IEEE Conf. on Computer Vision and Pattern
  Recognition}, 2018, pp. 6036--6046.

\bibitem{wen2017hyper-graph}
L.~Wen, Z.~Lei, M.-C. Chang, H.~Qi, and S.~Lyu, ``Multi-camera multi-target
  tracking with space-time-view hyper-graph,'' \emph{International Journal of
  Computer Vision}, vol. 122, no.~2, pp. 313--333, 2017.

\bibitem{le2018online}
Q.~C. Le, D.~Conte, and M.~Hidane, ``Online multiple view tracking: Targets
  association across cameras,'' in \emph{6th Workshop on Activity Monitoring by
  Multiple Distributed Sensing (AMMDS)}, 2018.

\bibitem{xu2016hierarchy}
Y.~Xu, X.~Liu, Y.~Liu, and S.-C. Zhu, ``Multi-view people tracking via
  hierarchical trajectory composition,'' in \emph{IEEE Conf. on Computer Vision
  and Pattern Recognition}, 2016, pp. 4256--4265.

\bibitem{olfati2007DKF}
R.~Olfati-Saber, ``Distributed kalman filtering for sensor networks,'' in
  \emph{IEEE Conf. on Decision and Control}, 2007, pp. 5492--5498.

\bibitem{soto2009distributed}
C.~Soto, B.~Song, and A.~K. Roy-Chowdhury, ``Distributed multi-target tracking
  in a self-configuring camera network,'' in \emph{IEEE Conf. on Computer
  Vision and Pattern Recognition}, 2009, pp. 1486--1493.

\bibitem{kamal2015ICF&JPDAF}
A.~T. Kamal, J.~H. Bappy, J.~A. Farrell, and A.~K. Roy-Chowdhury, ``Distributed
  multi-target tracking and data association in vision networks,'' \emph{IEEE
  Transactions on Pattern Analysis and Machine Intelligence}, vol.~38, no.~7,
  pp. 1397--1410, 2015.

\bibitem{kamal2012ICF}
A.~T. Kamal, J.~A. Farrell, and A.~K. Roy-Chowdhury, ``Information weighted
  consensus,'' in \emph{IEEE Conf. on Decision and Control (CDC)}, 2012, pp.
  2732--2737.

\bibitem{zhou1993JPDAF}
B.~Zhou and N.~Bose, ``Multitarget tracking in clutter: Fast algorithms for
  data association,'' \emph{IEEE Transactions on Aerospace and Electronic
  Systems}, vol.~29, no.~2, pp. 352--363, 1993.

\bibitem{he2019efficient}
L.~He, G.~Liu, G.~Tian, J.~Zhang, and Z.~Ji, ``Efficient multi-view
  multi-target tracking using a distributed camera network,'' \emph{IEEE
  Sensors Journal}, 2019.

\bibitem{shorinwa2020distributedVehicles}
O.~Shorinwa, J.~Yu, T.~Halsted, A.~Koufos, and M.~Schwager, ``Distributed
  multi-target tracking for autonomous vehicle fleets,'' in \emph{IEEE
  International Conference on Robotics and Automation}, 2020, pp. 3495--3501.

\bibitem{de2020divers}
K.~de~Langis and J.~Sattar, ``Realtime multi-diver tracking and
  re-identification for underwater human-robot collaboration,'' in \emph{IEEE
  Int. Conf. on Robotics and Automation}, 2020, pp. 11\,140--11\,146.

\bibitem{chandra2019densepeds}
R.~Chandra, U.~Bhattacharya, A.~Bera, and D.~Manocha, ``Densepeds: Pedestrian
  tracking in dense crowds using front-rvo and sparse features,'' in
  \emph{IEEE/RSJ Int. Conf. on Intelligent Robots and Systems}, 2019, pp.
  468--475.

\bibitem{virgona2018shoulder}
A.~Virgona, A.~Alempijevic, and T.~Vidal-Calleja, ``Socially constrained
  tracking in crowded environments using shoulder pose estimates,'' in
  \emph{IEEE Int. Conf. on Robotics and Automation}, 2018.

\bibitem{sharma2018cost}
S.~Sharma, J.~A. Ansari, J.~K. Murthy, and K.~M. Krishna, ``Beyond pixels:
  Leveraging geometry and shape cues for online multi-object tracking,'' in
  \emph{IEEE Int. Conf. on Robotics and Automation}, 2018, pp. 3508--3515.

\bibitem{tsai2019intertial}
R.~Y.-C. Tsai, H.~T.-Y. Ke, K.~C.-J. Lin, and Y.-C. Tseng, ``Enabling
  identity-aware tracking via fusion of visual and inertial features,'' in
  \emph{Int. Conf. on Robotics and Automation}, 2019, pp. 2260--2266.

\bibitem{chen2019abd}
T.~Chen, S.~Ding, J.~Xie, Y.~Yuan, W.~Chen, Y.~Yang, Z.~Ren, and Z.~Wang,
  ``Abd-net: Attentive but diverse person re-identification,'' in \emph{IEEE
  Int. Conf. on Computer Vision}, 2019, pp. 8351--8361.

\bibitem{quan2019auto}
R.~Quan, X.~Dong, Y.~Wu, L.~Zhu, and Y.~Yang, ``Auto-reid: Searching for a
  part-aware convnet for person re-identification,'' in \emph{IEEE Int. Conf.
  on Computer Vision}, 2019, pp. 3750--3759.

\bibitem{liu2018attributerecognition}
P.~Liu, X.~Liu, J.~Yan, and J.~Shao, ``Localization guided learning for
  pedestrian attribute recognition,'' in \emph{British Machine Vision
  Conference}, 2018.

\bibitem{zhou2019omni}
K.~Zhou, Y.~Yang, A.~Cavallaro, and T.~Xiang, ``Omni-scale feature learning for
  person re-identification,'' in \emph{IEEE Int. Conf. on Computer Vision},
  2019, pp. 3702--3712.

\bibitem{wu2019detectron2}
Y.~Wu, A.~Kirillov, F.~Massa, W.-Y. Lo, and R.~Girshick, ``Detectron2,''
  \url{https://github.com/facebookresearch/detectron2}, 2019.

\bibitem{he2017event}
X.~He, C.~Hu, W.~Xue, and H.~Fang, ``On event-based distributed kalman filter
  with information matrix triggers,'' \emph{IFAC-PapersOnLine}, vol.~50, no.~1,
  pp. 14\,308--14\,313, 2017.

\bibitem{wei2018msmt17}
L.~Wei, S.~Zhang, W.~Gao, and Q.~Tian, ``Person transfer gan to bridge domain
  gap for person re-identification,'' in \emph{IEEE Conf. on Computer Vision
  and Pattern Recognition}, 2018, pp. 79--88.

\bibitem{k-short}
J.~Berclaz, F.~Fleuret, E.~Turetken, and P.~Fua, ``Multiple object tracking
  using k-shortest paths optimization,'' \emph{IEEE Transactions on Pattern
  Analysis and Machine Intelligence}, vol.~33, no.~9, pp. 1806--1819, 2011.

\bibitem{bernardin2008evaluating}
K.~Bernardin and R.~Stiefelhagen, ``Evaluating multiple object tracking
  performance: the clear mot metrics,'' \emph{EURASIP Journal on Image and
  Video Processing}, vol. 2008, pp. 1--10, 2008.

\bibitem{ristani2016evaluating2}
E.~Ristani, F.~Solera, R.~Zou, R.~Cucchiara, and C.~Tomasi, ``Performance
  measures and a data set for multi-target, multi-camera tracking,'' in
  \emph{European Conference on Computer Vision}, 2016, pp. 17--35.

\bibitem{pom}
F.~Fleuret, J.~Berclaz, R.~Lengagne, and P.~Fua, ``Multicamera people tracking
  with a probabilistic occupancy map,'' \emph{IEEE Transactions on Pattern
  Analysis and Machine Intelligence}, vol.~30, no.~2, pp. 267--282, 2007.

\end{thebibliography}
}
\end{document}